\documentclass[10pt,twocolumn,letterpaper]{article}

\usepackage{cvpr}              

\usepackage{graphicx}
\usepackage{amsmath,bm}
\usepackage{amssymb}
\usepackage{booktabs}
\usepackage[accsupp]{axessibility} 
\DeclareMathOperator*{\argmin}{\arg\min} 

\usepackage{enumitem}

\usepackage[pagebackref,breaklinks,colorlinks]{hyperref}

\usepackage[capitalize]{cleveref}
\crefname{section}{Sec.}{Secs.}
\Crefname{section}{Section}{Sections}
\Crefname{table}{Table}{Tables}
\crefname{table}{Tab.}{Tabs.}

\def\cvprPaperID{6664} 
\def\confName{CVPR}
\def\confYear{2023}

\makeatletter
\DeclareRobustCommand\onedot{\futurelet\@let@token\@onedot}
\def\@onedot{\ifx\@let@token.\else.\null\fi\xspace}
\def\eg{\emph{e.g}\onedot} 
\def\ie{\emph{i.e}\onedot}

\makeatother

\DeclareRobustCommand*{\IEEEauthorrefmark}[1]{%
    \raisebox{0pt}[0pt][0pt]{\textsuperscript{\footnotesize\ensuremath{#1}}}}

\begin{document}

\title{Noisy Correspondence Learning with Meta Similarity Correction}
\author{Haochen Han\IEEEauthorrefmark1,
Kaiyao Miao\IEEEauthorrefmark2,
Qinghua Zheng\IEEEauthorrefmark1, 
Minnan Luo\IEEEauthorrefmark{1}\thanks{Minnan Luo is the corresponding author.} 
\and
{\normalsize \IEEEauthorrefmark1 School of Computer Science and Technology, Xi’an Jiaotong University, China}
\and
{\normalsize \IEEEauthorrefmark2 School of Cyber Science and Engineering, Xi’an Jiaotong University, China}
\and
\texttt{\small \{hhc1997, miaoky814\}@stu.xjtu.edu.cn, \{qhzheng, minnluo\}@xjtu.edu.cn}}

\maketitle

\begin{abstract}
   Despite the success of multimodal learning in cross-modal retrieval task, the remarkable progress relies on the correct correspondence among multimedia data. However, collecting such ideal data is expensive and time-consuming. In practice, most widely used datasets are harvested from the Internet and inevitably contain mismatched pairs. Training on such noisy correspondence datasets causes performance degradation because the cross-modal retrieval methods can wrongly enforce the mismatched data to be similar. To tackle this problem, we propose a Meta Similarity Correction Network (MSCN) to provide reliable similarity scores. We view a binary classification task as the meta-process that encourages the MSCN to learn discrimination from positive and negative meta-data. To further alleviate the influence of noise, we design an effective data purification strategy using meta-data as prior knowledge to remove the noisy samples. Extensive experiments are conducted to demonstrate the strengths of our method in both synthetic and real-world noises, including Flickr30K, MS-COCO, and Conceptual Captions. Our code is publicly available.\footnote{\url{https://github.com/hhc1997/MSCN}}
   
\end{abstract}

\section{Introduction}
Recently, cross-modal retrieval has drawn much attention with the rapid growth of multimedia data. Given a query sample of specific modality, cross-modal retrieval aims to retrieve relevant samples across different modalities. Existing cross-modal retrieval works \cite{arandjelovic2018objects,deng2018triplet,guo2019image} usually learn a comparable common space to bridge different modalities, which achieved remarkable progress in many applications, including video-audio retrieval \cite{arandjelovic2018objects,suris2018cross}, visual question answering \cite{ma2016learning,guo2019image}, and image-text matching \cite{zhang2020context}.

\begin{figure}[t]
\centering  
\includegraphics[width=\columnwidth]{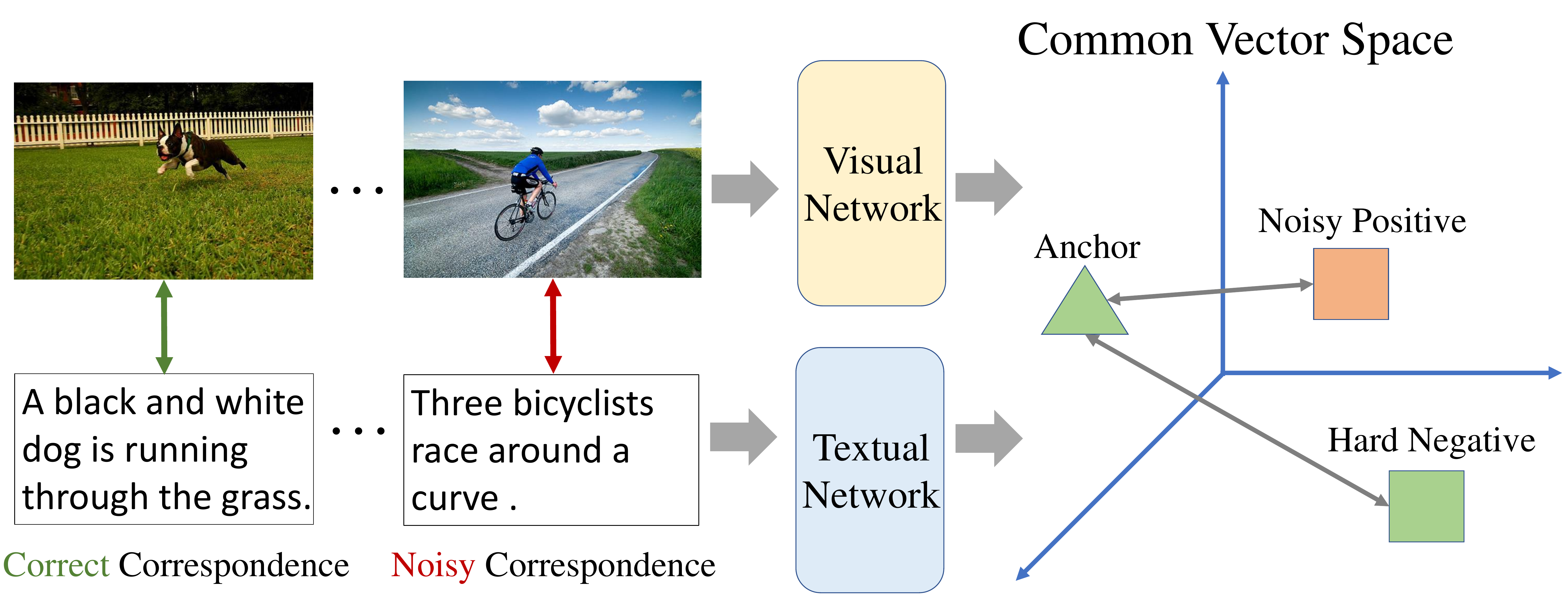}
\caption{Illustration of noisy correspondence in image-text retrieval. A standard triplet loss is used to enforce the positive pairs to be closer than negatives in the common space. Noisy correspondence is the mismatched pairs but wrongly considered as positive ones, and thus results in model performance degradation.  }
\label{fig_intro}
\vspace{-0.2cm} 
\end{figure}

Despite the promise, a core assumption in cross-modal retrieval is the correct correspondence among multiple modalities. However, collecting such ideal data is expensive and time-consuming. In practice, most widely used datasets are harvested from the Internet and inevitably contain noisy correspondence \cite{sharma2018conceptual}. As illustrated in \cref{fig_intro}, the cross-modal retrieval method will wrongly enforce the mismatched data to be similar when learning with noisy correspondence, which may significantly affect the retrieval performance. To date, few effort has been made to address this. Huang \cite{huang2021learning} first researches this issue and proposes the NCR method to train from the noisy image-text pairs robustly. Inspired by the prior success for noisy labels \cite{li2019dividemix}, NCR divides the data into clean and noisy partitions and rectifies the correspondence with an adaptive model. However, NCR based on the memorization effect of DNNs \cite{arpit2017closer}, which leads to poor performance under high noise ratio.

To tackle the challenge, we propose a \textbf{Meta Similarity Correction Network (MSCN)} which aims to provide reliable similarity scores for the noisy features from main net. We view a binary classification task as the meta-process: given a multimodal sample, the MSCN will learn to determine whether the modalities correspond to each other or not, where the prediction of MSCN can be naturally regarded as the similarity score. Specifically, a small amount of clean pairs is used to construct positive and negative meta-data, both viewed as meta-knowledge that encourages MSCN to learn the discrimination. Meanwhile, the main net trained by the corrected similarity score from MSCN with a self-adaptive margin to achieve robust learning. This interdependency makes the main net and MSCN benefits each other against noise. However, due to the property of triplet loss, it remains to produce positive loss for noisy pairs even if we employ the ideal similarity scores. To this end, we further propose a meta-knowledge guided data purification strategy to remove samples with potentially wrong correspondence. Extensive experiments are conducted to demonstrate the strengths of our method in both synthetic and real-world noises.

The main contributions of this work are summarized as follows:

\setlist{nolistsep}
\begin{itemize}[noitemsep]
    \item We pioneer the exploration of meta-learning for noisy correspondence problem, where a meta correction network is proposed to provide reliable similarity scores against noise.
    
    
    \item  We present a novel meta-process that first considers both positive and negative data as meta-knowledge to encourage the MSCN to learn discrimination from them.
    
    \item We design an effective data purification strategy using meta-data as prior knowledge to remove the noisy samples.
    
\end{itemize}

\section{Related Work}
\label{sec:related work}
\subsection{Cross-Modal Retrieval}
Cross-modal retrieval works aim to take one modality of data as query to retrieve the related data in other modalities. Based on the utilization of annotation information, cross-modal retrieval methods can be roughly divided into two categories: 1) Unsupervised Methods\cite{vse++,liu2020graph,wang2016learning,li2019visual}. It only uses the co-occurrence information to learn the common representations among the multimodal data. 2) Supervised methods \cite{tang2016supervised,zhen2019deep,xu2019deep,hu2021learning}. The extra label information is employed to boost the discrimination of common representations. However, the lack of annotations limited the practicability in the real world.

Our proposed approach falls in the category of unsupervised methods. For example, Wang \cite{wang2016learning} uses a two-branch neural networks to learn the joint embeddings of multimodal data.
Inspired by hard negative mining, VSE++ \cite{vse++} uses hard negatives to improve the retrieval performance. To capture fine-grained interplay between modalities, SCAN \cite{lee2018stacked} introduce stacked cross attention to enable attention with context from both image and sentence. Recently, motivated by the powerful learning ability of graph model, GSMN \cite{liu2020graph} and SGRAF \cite{diao2021similarity} construct graph structure for multimodal data to benefit the learning of fine-grained correspondence. Although existing works have achieved remarkable results, they usually depend on the correct correspondence among multimodal data and cannot tackle the noise issue. Thus, it is significant to explore how to learn cross-modal retrieval with noisy correspondence, but which is rarely touched in previous studies.

\subsection{Noisy Correspondence Learning}
As a newly proposed problem, noisy correspondence is the mismatched pairs but wrongly considered as positive ones. Huang \cite{huang2021learning} first research this issue and propose the NCR method to robustly train the image-text matching model with noisy correspondence pairs. Inspired by the prior success for noisy labels \cite{li2019dividemix}, NCR divides the data into clean and noisy part based on the memorization effect of DNNs, and then rectifies the correspondence with an adaptive model. Qin\cite{qin2022deep} proposes a uncertainty-based method to achieve efficient learning. Moreover, some works \cite{yang2021partially,yang2022robust} study the partially view-unaligned problem which can be viewed as a generalized noisy correspondence learning. Recently, some works \cite{yang2022learning,han2022noise} explore a more challenging scenario where noisy labels and noisy correspondence may occur simultaneously. In contrast, our work tackles this challenge in cross-modal retrieval from a meta-learning perspective.


\begin{figure*}[!t]
  \centering
  \includegraphics[width=1\textwidth]{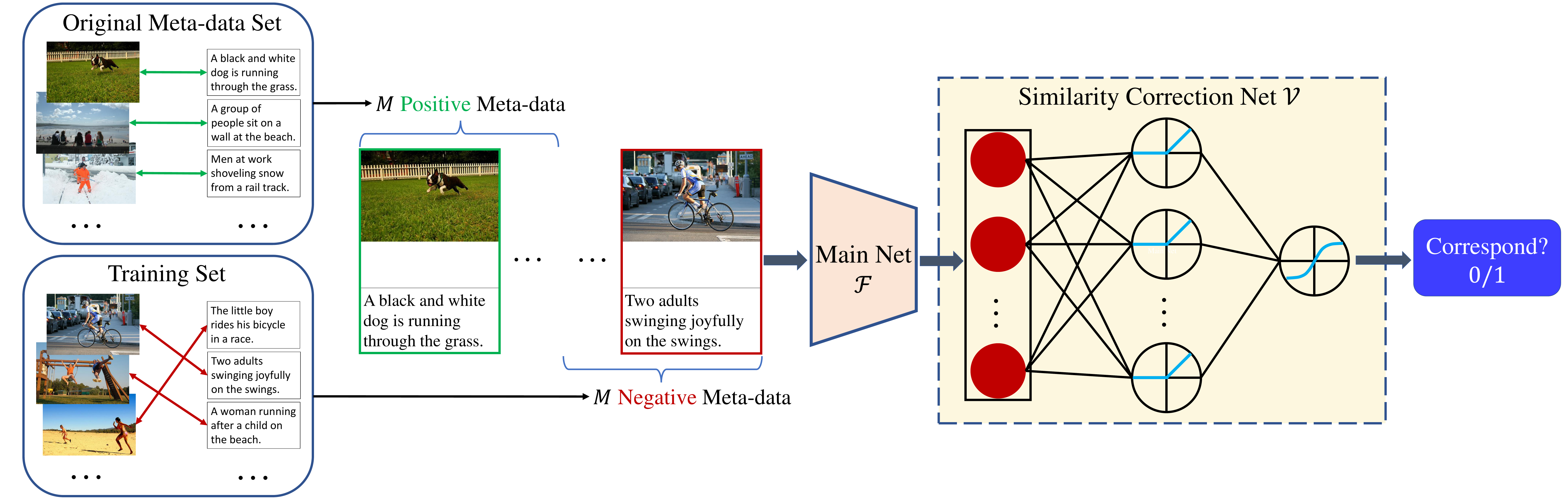}
  \caption{Overview of the proposed meta-process. The positive pairs are taken from the original meta-data which contain corresponding image and text, while negative pairs are extracting from different pairs in training set to construct mismatched pairs. Then the MSCN is learned to determine whether the modalities correspond to each other or not.} \label{fig:meta-process}
\vspace{-0.2cm} 
\end{figure*}

\subsection{Meta-Learning}
The objective of meta-learning is to learn at a more advanced level than conventional learning, such as learning the update rule \cite{mishra2018simple}, finding easily fine-tuned parameters \cite{finn2017model}, or adapting to new tasks \cite{li2018learning}. Recently, researchers use meta-learning to find model's parameters that robust against noisy labels \cite{li2019learning,shu2019meta,zheng2021meta}. For example, Li \cite{li2019learning} optimizes a meta-objective before conventional training to enable the model not overfit to noise. MW-Net \cite{shu2019meta} use a meta-process to automatically assign weights to the training samples. MLC \cite{zheng2021meta} presents a label correction network trained as a meta manner which generate reliable labels for the noisy training data.

The most relevant existing method is MLC \cite{zheng2021meta}, which first poses the noisy label problem as a meta label correction. However, we argue that our proposed method differs from MLC in two aspects. First, we focuses on the noisy correspondence problem in multimodal data instead of the unimodal classification scenarios. Second, we leverage both positive and negative meta-data to guide model learning the discrimination, which is unexplored in previous meta-learning methods.

\section{The Proposed Method}
\label{sec:method}
\subsection{Problem Formulation}
Without losing generality, we use the image-text retrieval task as a proxy to investigate the noisy correspondence problem in cross-modal retrieval. Given a training set $\mathcal{D}_{train}=\{(I_i,T_i,y_i) \}_{i=1}^N$, where $(I_i,T_i)$ is the $i$-th image-text pair, $y_i\in \{1,0\}$ indicates the pair is positive (matched), \ie $(I_i,T_i)$, or negative (mismatched), \ie $(I_i,T_{j\neq i})$, and $N$ is the number of the entire training data. The noisy correspondence in bimodal data is the negative pair but wrongly considered as $y_i=1$. To address this issue, we propose a meta-learning based method to achieve robust training. The details are delineated next.

\subsection{Meta-Learning Based Similarity Correction}
\paragraph{The Training Objective.} To begin with, we describe the standard projection module to enable cross-modal features to be comparable. Let $f(I;W_f)$ and $g(T;W_g)$ be the modal-specific networks to map the visual and textual modalities into the joint embedding space, respectively. Then we compute the similarity feature between bimodal features by the function 
\begin{equation}  
    \label{eq:similarity function}
        S\left(f\left(I\right),g\left(T\right);W_s\right) = \frac{W_s \lvert f(I) - g(T)\rvert^2 }{\lVert W_s \lvert f(I) - g(T)\rvert^2\rVert_2},
\end{equation}
where $W_s$ is a learnable parameter matrix to obtain low-dimensional similarity representation. For notation convenience, we collect $\{W_f,W_g,W_s\}$ into $W$ as the parameters of main net $\mathcal{F}=(f,g,S)$ and denote similarity feature as $\mathcal{F}_W(I,T)$. With the presence of noisy correspondence, the model will inevitably overfit to noise if we use the similarity representation directly. To tackle this problem, we propose a meta similarity correction network (MSCN), regarding process of obtaining similarity score as a meta-process, which takes the noisy similarity representation as input and produces the corrected similarity score. Specifically, our MSCN is formulated as a MLP network $\mathcal{V}_{\Theta}(\cdot)$ with parameters $\Theta$. The output $s_i = \mathcal{V}_{\Theta}(\mathcal{F}_W(I_i,T_i))$ is activated with Sigmoid function to represent the similarity score located in $[0,1]$. 

\begin{figure*}[!t]
  \centering
  \includegraphics[width=1\textwidth]{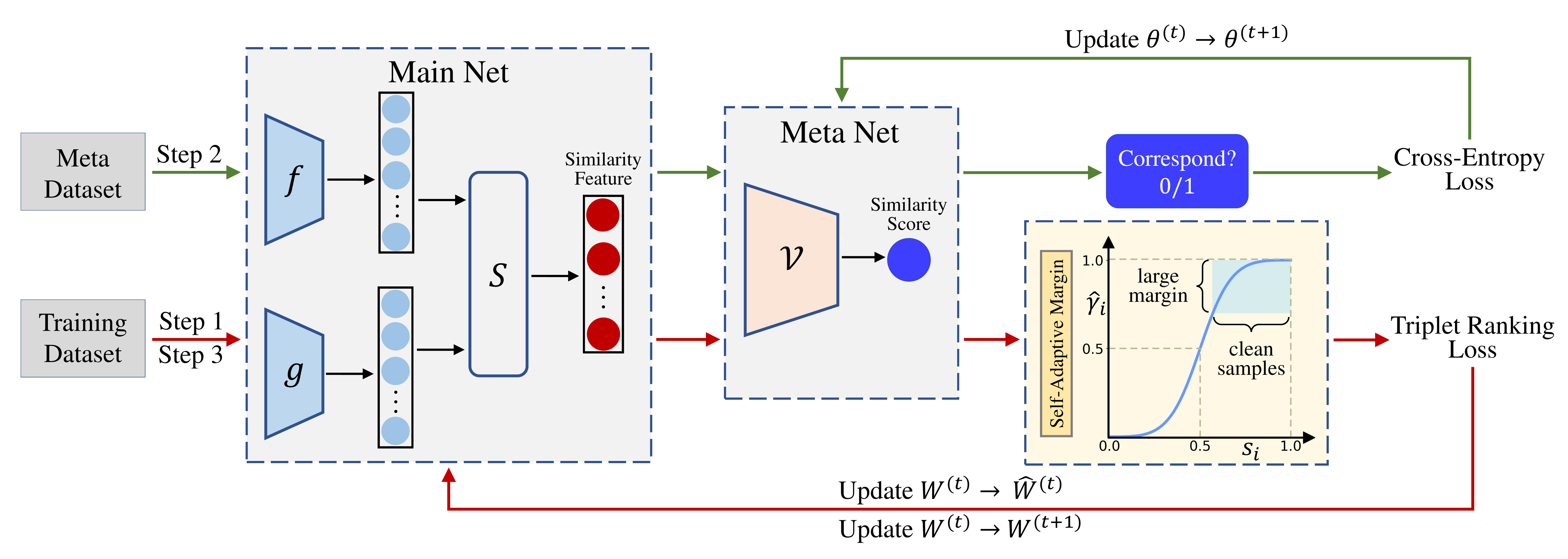}
  \caption{Flowchart of the proposed bi-level optimization. At each iteration, we alternately update the main net and the meta net (MSCN) following the steps in Eq. \eqref{eq:update W_t} - \eqref{eq:update W_t+1}.} \label{fig:meta-update}
 \vspace{-0.2cm} 
\end{figure*}

For the noisy training pairs, the main net learns noise-tolerant features guided by the corrected similarity scores produced from MSCN; while the MSCN predicts the reliable similarity score based on features from main net. This interdependency enables the main net and meta net reinforce each other against noise, and optimized via a bi-level optimization. Formally, the optimal parameters $W^*$ is calculated by minimizing the cumulative loss over training data:

\begin{equation}  
    \label{eq:optimal W}
    W^*(\Theta) = \argmin_{W} \mathbb{E}_{(I_i,T_i)\in \mathcal{D}_{train}} \  l^{train}(I_i,T_i),
\end{equation}
where $l^{train}(\cdot)$ is the triplet ranking loss:

{\small
\begin{equation}
\begin{aligned}
    \label{eq:triplet loss}
    l^{train}(I_i,T_i) &=[\hat{\gamma} - \mathcal{V}_{\Theta}\left (\mathcal{F}_W(I_i,T_i) \right) + \mathcal{V}_{\Theta}\left (\mathcal{F}_W(I_i,T_i^-) \right)]_+
    \\ &+ [\hat{\gamma} - \mathcal{V}_{\Theta}\left (\mathcal{F}_W(I_i,T_i) \right) + \mathcal{V}_{\Theta}\left (\mathcal{F}_W(I_i^-,T_i) \right)]_+,
\end{aligned}
\end{equation}}
where $[x]_+ = max(x,0)$, $I^-_i$ and $T^-_i$ are the hardest negatives corresponding to given pair $(I_i,T_i)$ similar to VSE++\cite{vse++}. To achieve robust cross-modal retrieval, we desigin a self-adaptive margin $\hat{\gamma}$ instead of a fixed value in VSE++. Ideally, the margin should put less value on the noisy pairs and more on the clean pairs. To this end, we adjust the margin according to the similarity score:
 \begin{equation}
    \label{eq: adaptive margin}
        \hat{\gamma} = \frac{1}{1+(\frac{s_i}{1-s_i})^{-\tau}}\gamma,
\end{equation}
where $\gamma$ is the original margin value and $\tau >0 $ is the parameter. This self-adaptive margin will enable the triplet loss put more attention on the samples with high similarity scores that possibly are clean pairs.

\paragraph{Meta Training Process.} To ensure the reliability of the similarity scores produced by MSCN, we utilize a small amount clean meta-data set $\mathcal{D}_{meta}=\{(I_i,T_i) \}_{i=1}^M$ to guide the training, where $M$ is much smaller than $N$. While meta-learning has been extensively studied in recent literature \cite{franceschi2018bilevel,shu2019meta,zheng2021meta}, it remains challenging to construct the meta-process in our objective. On the one hand, the similarity features from main net can not used directly to produce loss. On the other hand, the meta-process should enable the output of MSCN to represent the similarity score. To achieve this, we view a binary classification task as the meta-process: given an image-text pair, the MSCN is learned to determine whether the modalities correspond to each other or not. Specifically, the positive pairs are taken from the meta-data which contain corresponding image and text, \ie $(I_i,T_i,y_i=1)$, while negative pairs are extracting from different pairs in training set to construct mismatched pairs, \ie $(I_i,T_{j \neq i},y_i=0)$. The binary labels are produced from the data itself, and can be regarded as the ideal similarity score to guide the training of MSCN. The prediction of MSCN measures the probability of a pair being clean which can be naturally equivalent to the similarity score. Formally, in each iteration of training, we randomly construct $M$ mismatched pairs from training set to extend the meta-data set as $\mathcal{D}_{meta}^{\prime}=\{(I_i,T_i,y_i) \}_{i=1}^{2M}$. Then the optimal parameter $\Theta ^*$ can be learned by minimizing the following meta-loss calculated on meta-data:

\begin{equation}  
\begin{aligned}
    \label{eq:optimal Θ}
    \Theta ^* &= \argmin_{\Theta} \mathbb{E}_{(I_i,T_i)\in \mathcal{D}_{meta}^{\prime}} \  l^{meta}(I_i,T_i,y_i),
\end{aligned}
\end{equation}
where $l^{meta}(\cdot)$ is cross-entropy loss \cite{wei2016cross,wang2017adversarial}:
\begin{equation} 
    \label{eq: cross-entropy loss}
    l^{meta}(I_i,T_i,y_i)= -y_i \cdot log\mathcal{V}_{\Theta} \left(\mathcal{F}_{W^*(\Theta)}\left(I_i,T_i \right)\right).
\end{equation}
As illustrated in \cref{fig:meta-process}, despite its simplicity, our proposed meta-process can not only enforce the MSCN to learn reliable similarity score but also has the following advantages: 1) previous works only guide the meta net with ground truth data; however, we leverage both positive and negative data as meta-knowledge that encourage the MSCN to learn discrimination from them. 2) We overcome one limitation of meta-learning methods that the lack of meta-data. We extract from different pairs among training set to construct massive negative meta-data. Moreover, it is more suitable to the rule of noisy correspondence, which the irregularity of noisy data causes in various patterns.

\paragraph{Bi-level Optimization.} Motivated by recent works \cite{finn2017model,ren2018learning,shu2019meta,ji2021meta}, we use a bi-level optimization strategy to guarantee the efficiency for updating our main net and MSCN. As illustrated in \cref{fig:meta-update}, it is practically solved via two nested loops of optimization. Specifically, at each iteration $t$ of training, we sample a mini-batch of training pairs $\{(I_i,T_i)\}_{i=1}^n$, where $n$ is the size of mini-batch. We first update the main net parameters by taking a gradient descent step toward the direction of triplet loss, which can be formulated as follows:
\begin{equation}
    \label{eq:update W_t}
    	\hat{W}^{(t)}(\Theta) = W^{(t)} - \alpha \sum_{i=1}^{n} \nabla_W l^{train}(I_i,T_i)\big|_{W^{(t)}},
\end{equation}
where $\alpha$ means the learning rate. The updated $\hat{W}^{(t)}$ can be regard as the approximation of the optimal $W^*(\Theta)$ at $t$ iteration, containing the gradient information of $\Theta$. Then we update the parameters of MSCN with main net parameters fixed. Guided by the objective loss in Eq. \eqref{eq:optimal Θ} and a mini-batch of meta-data $\{(I_i,T_i,y_i)\}_{i=1}^m$, where $m$ is the mini-batch size. The updated parameters of MSCN is calculated as:
\begin{equation}
    \label{eq:update Θ_t}
    	\Theta^{(t+1)} = \Theta^{(t)} - \beta \frac{1}{m} \sum_{i=1}^{m} \nabla_{\Theta}  l^{meta}(I_i,T_i,y_i) \big|_{\Theta^{(t)}},
\end{equation}
where $\beta$ is the learning rate for meta-process. Finally, we update the parameters of main net while fixing the MSCN parameters. With the corrected similarity scores produced by the newly updated $\Theta^{(t+1)}$, we can update the main net parameters $W^{(t+1)}$ by employing the gradient decent as:
\begin{equation}
    \label{eq:update W_t+1}
    	W^{(t+1)} = W^{(t)} - \alpha \sum_{i=1}^{n} \nabla_W l^{train}(I_i,T_i)\big|_{W^{(t)}}.
\end{equation}

The above procedure proceeds in an alternating manner to optimize the main net and MSCN until it exceeds the maximal iteration.

\subsection{Meta-Knowledge Guided Data Purification}
\begin{figure}[t]
\centering  
\includegraphics[width=1\columnwidth,trim=35 5 50 35,clip]{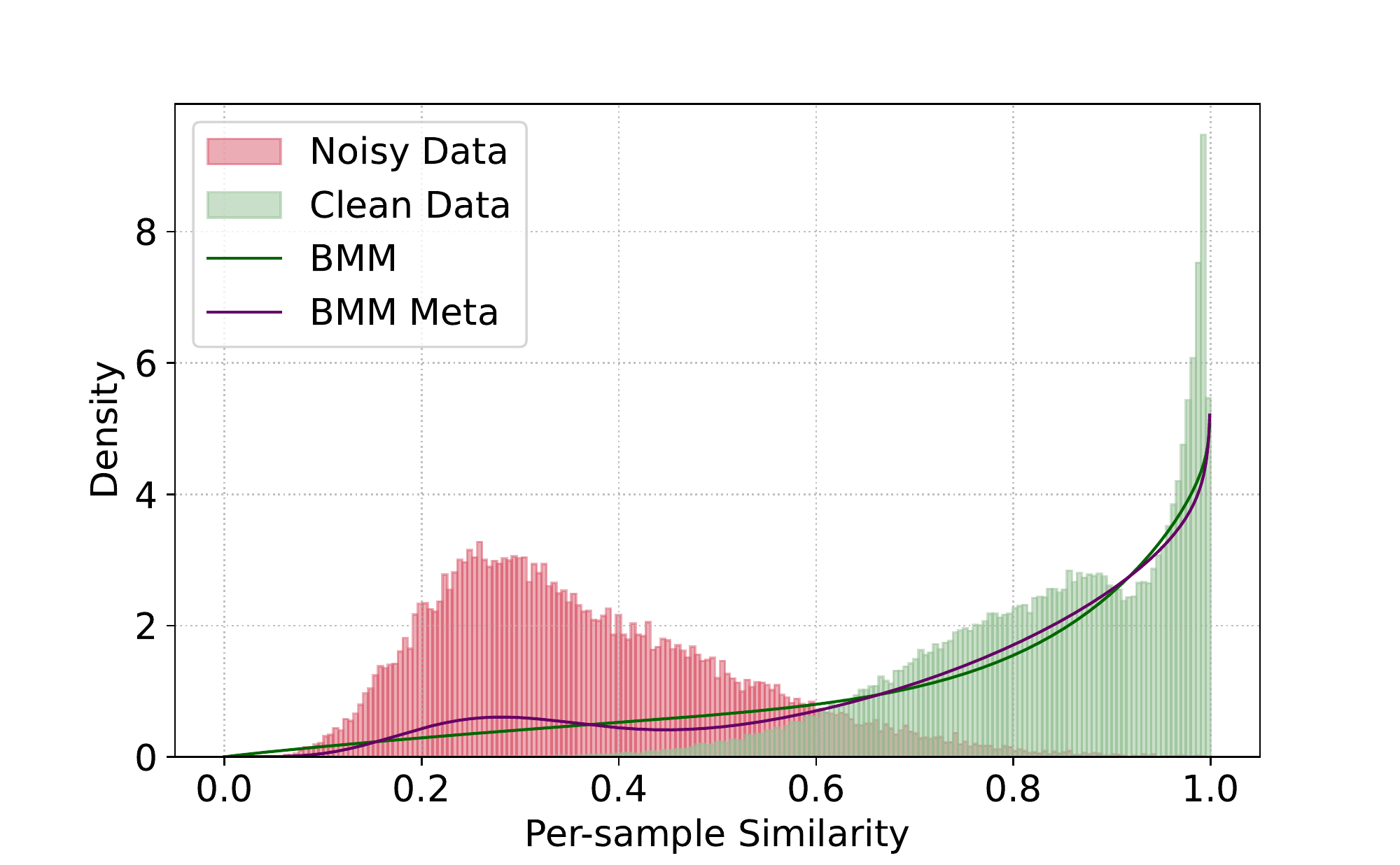}
\caption{Per-sample similarity score distribution, estimated BMM and BMM with meta prior knowledge under 20\% noisy correspondence in Flickr30K after 5 epochs warmup.}
\label{BMM}
\vspace{-0.1cm} 
\end{figure}

As mentioned, our MSCN can provide corrected similarity score to avoid noise. However, due to the property of triplet loss, it remains to produce positive loss for the noisy pairs even if we employ the ideal similarity score (\eg, $ \mathcal{V}\left (F(I_i,T_i) \right) 	\approx 0 < \mathcal{V}\left (F(I_i^-,T_i) \right) + \hat{\gamma}$), which results in the model fitting to the noise. To tackle this issue, we adopt sample selection strategy to remove the noisy data and only use clean samples as training data. Previous works \cite{arazo2019unsupervised,yao2020searching} motivated by the memorization effect of DNNs that clean sample has a lower loss during the beginning of training. Unfortunately, the triplet ranking loss is more complex, \ie, affected by both positive and negative samples, which leads to sub-optimal selection results. Based on the observation shown in \cref{BMM}, the clean and noisy pairs are prone to distinguish from the similarity scores. Gaussian Mixture Model (GMM) is a widely used technique to model mixture distribution \cite{li2019dividemix,huang2021learning}. However, we find the similarity distribution exhibits high skew toward $1$ and causes a poor approximation for GMM. Therefore, we fit a two-component Beta Mixture Model (BMM) \cite{ma2011bayesian} of per-pair similarity score $s_i$ to better distinguish clean and noisy samples:

\begin{equation}
    \label{eq: two-component BMM}
    	p(s_i) = \sum_{k=1}^{K} \lambda_k \phi(s_i|\alpha_k,\beta_k),
\end{equation}
where $K=2$ and $\lambda_k$ is the mixture coefficient, and $\phi(s|\alpha_k,\beta_k)$ denotes the probability density function with parameters $\alpha_k,\beta_k > 0$.

We use the Expectation-Maximization \cite{dempster1977maximum} algorithm to optimize the BMM, which poses a crucial challenge: the initialization of model parameters. Existing methods typically adopt the $K$-Means algorithm \cite{likas2003global} for initializing; however, it is time-consuming (\ie, the time complexity is $O(TKN)$, where $T$ is the number of iterations) and underperforming. We thus propose a novel initialization approach which takes the meta-data as prior knowledge. Specifically, we fit the two Beta components with positive and negative meta-data to model the distribution of clean and noisy pairs, respectively. Given the similarity scores $\mathcal{S}_{p} = \{ s_i \}_{i=1}^{M}$ corresponding to positive meta-data, the parameters of the component modeling clean distribution can be initialled by
\begin{equation}
\begin{aligned}
    \label{eq: initialize alpha}
    	 \alpha_k &= \frac{\left(1-E\left(\mathcal{S}_{p}\right)\right){E(\mathcal{S}_{p})}^2}{V(\mathcal{S}_{p})} - E(\mathcal{S}_{p}),
        \\ \beta_k  &=  \frac{\alpha_k \left(1-E\left(\mathcal{S}_{p}\right)\right)}{E(\mathcal{S}_{p})},
\end{aligned}
\end{equation}
where $E(\cdot)$ and $V(\cdot)$ are the mean and variance function, respectively. The initialization of component modeling noisy distribution is analogous but uses similarity scores $\mathcal{S}_{n} = \{ s_i \}_{i=1}^{M}$ of negative meta-data. As illustrated in \cref{BMM}, our meta-knowledge guided initialization method achieves better approximation while with lower time complexity (\ie, $O(KM)$).

Finally, we compute the posterior probability $p_i=p(k|s_i) = p(k)p(s_i|\alpha_k,\beta_k)/p(s_i)$ as the probability of $i$-th simple being clean. The purified training set is defined as:
 \begin{equation}
    \label{eq: selected set}
        \mathcal{D}_{train}^{\prime} = \{(I_i,T_i) \in \mathcal{D}_{train} |  p_i > 0.5\}.
\end{equation}
At each iteration, we produce the purified dataset before the training of our meta-learning objective. Moreover, following  \cite{li2019dividemix,huang2021learning,han2018co}, we maintain two networks $\{\mathcal{F}_W^1,\mathcal{V}_{\Theta}^1\}$ and $\{\mathcal{F}_W^2,\mathcal{V}_{\Theta}^2\}$ to avoid error accumulation that one network will produce the purified training set to train the other one. The full algorithm 
is shown in \emph{supplementary material}.

\section{Experiment}
\label{sec:experiment}

\subsection{Datasets and Evaluation Protocol}
\paragraph{Datasets.} We evaluate our method on three standard image-text retrieval datasets. To be specific, the Flickr30K \cite{young2014image} collects 31,783 images with 5 corresponding captions each from Flickr website. Following the split in \cite{huang2021learning}, we use 1,000 image-text pairs for validation, 1,000 image-text pairs for testing, and the rest for training. The MS-COCO \cite{lin2014microsoft} contains 123,287 images with 5 corresponding captions each. We split the MS-COCO following \cite{huang2021learning} that 113,287 image-text pairs for training, 5,000 pairs for validation, and 5,000 pairs for testing. Conceptual Captions \cite{sharma2018conceptual} is a large-scale dataset contains 3.3M images with one corresponding caption each. We employ the same subset following \cite{huang2021learning}, \ie, CC152K. Specifically, CC152K consists of 150,000 pairs from training set for training, 1,000 pairs from validation set for validation, and 1,000 pairs from validation set for testing. Moreover, we use an extra meta-data set from validation containing 3,000 pairs.

\paragraph{Evaluation Protocol.} We evaluate the retrieval performance with the recall rate at K (R@K) metric. In a nutshell, R@K is the proportion for the relevant items retrieved in the closest K items to the query. In our experiments, we take image and text as queries, respectively, and report R@1, R@5, and R@10 results for a comprehensive evaluation. Following \cite{huang2021learning}, we average the similarity scores from two networks at the inference phase.

\subsection{Implementation Details}
As a general approach, our method can be easily applied to almost all cross-modal retrieval methods to improve robustness . For fair comparison, we chose the same network backbones with NCR \cite{huang2021learning}, \ie, the projection modules and similarity function. For all datasets, the training processes contain 50 epochs after a 5 epochs warmup. We employ ADAM \cite{kingma2015adam} as the optimizer for both main net and meta net with a batch size of 64. To fit the BMM, we set the stop threshold as $10^{-2}$ and maximum number of iterations as 10 for the EM procedure. Note that, we clamp the similarity scores into a range of $[10^{-4},1-10^{-4}]$ instead of [0,1] for stable fitting. Moreover, we set the margin parameter as $0.2$ and $\tau = 2$ to calculate the self-adaptive margin. For all datasets, the number of ground-truth meta-data is approximately $2\%$ of training data, and the learning rate of main net and meta net are initialized with $2\times 10^{-4}$ and $1.7 \times 10^{-5}$, respectively. For CC152K containing real noise, we employ the validation data as the meta set due to the lack of ground-truth data. For all experiments, we decay the learning rate by $0.1$ after 30 epochs.

\begin{table}[h]
  \setlength {\belowcaptionskip} {-0.4cm}
  \centering
    \renewcommand\arraystretch{1}
    \begin{tabular}{c|cccc}
    \toprule
          & \multicolumn{4}{c}{Image to Text} \\
    \midrule
    Methods & R@1   & R@5   & R@10  & SUM \\
    \midrule
    SCAN (ECCV’18) & 30.5  & 55.3  & 65.3  & 151.1  \\
    VSRN(ICCV’19) & 32.6  & 61.3  & 70.5  & 164.4  \\
    IMRAM(CVPR’20) & 33.1  & 57.6  & 68.1  & 158.8  \\
    SAF (AAAI’21) & 31.7  & 59.3  & 68.2  & 159.2  \\
    SGR (AAAI’21) & 11.3  & 29.7  & 39.6  & 80.6  \\
    SGR* (AAAI’21) & 35.0  & 63.4  & 73.3  & 171.7  \\
    NCR(NIPS’21) & 39.5  & 64.5  & 73.5  & 177.5  \\
    MSCN*(Ours) & 39.7  & 65.4  & 75.3  & 180.4  \\
    MSCN(Ours) & \textbf{\ 40.1 } & \textbf{\ 65.7 } & \textbf{\ 76.6 } & \textbf{\ 182.4 } \\
    \midrule
          & \multicolumn{4}{c}{Text to Image} \\
    SCAN (ECCV’18) & 26.9  & 53.0  & 64.7  & 144.6  \\
    VSRN(ICCV’19) & 32.5  & 59.4  & 70.4  & 162.3  \\
    IMRAM(CVPR’20) & 29.0  & 56.8  & 67.4  & 153.2  \\
    SAF (AAAI’21) & 31.9  & 59.0  & 67.9  & 158.8  \\
    SGR (AAAI’21) & 13.1  & 30.1  & 41.6  & 84.8  \\
    SGR* (AAAI’21) & 34.9  & 63.0  & 72.8  & 170.7  \\
    NCR(NIPS’21) & 40.3 & 64.6  & 73.2  & 178.1  \\
    MSCN*(Ours) & 39.8  & 66.1  & 75.0  & 180.9  \\
    MSCN(Ours) &\textbf{\ 40.6 }  & \textbf{\ 67.4 } & \textbf{\ 76.3 } & \textbf{\ 184.3 } \\
    \bottomrule
    \end{tabular}%
  \caption{Image-text retrieval performance on CC152K with real noise, and the best results are highlighted in \textbf{bold}.}
  \label{tab:table1}%
\end{table}%

\begin{table*}[htbp]
  \setlength {\belowcaptionskip} {-0.4cm}
  \centering
  \scalebox{1}{
    \renewcommand\arraystretch{0.97}
    \small
    \begin{tabular}{c|c|ccc|ccc|ccc|ccc}
    \toprule
          &       & \multicolumn{6}{c|}{Flickr30K}                & \multicolumn{6}{c}{MS-COCO} \\
          &       & \multicolumn{3}{c|}{Image to Text} & \multicolumn{3}{c|}{Text to Image} & \multicolumn{3}{c|}{Image to Text} & \multicolumn{3}{c}{Text to Image} \\
    \midrule
    Noise  & Methods & R@1   & R@5   & R@10  & R@1   & R@5   & R@10  & R@1   & R@5   & R@10  & R@1   & R@5   & R@10 \\
    \midrule
          & SCAN  & 59.1  & 83.4  & 90.4  & 36.6  & 67.0  & 77.5  & 66.2  & 91.0  & 96.4  & 45.0  & 80.2  & 89.3  \\
          & VSRN  & 58.1  & 82.6  & 89.3  & 40.7  & 68.7  & 78.2  & 25.1  & 59.0  & 74.8  & 17.6  & 49.0  & 64.1  \\
          & IMRAM & 63.0  & 86.0  & 91.3  & 41.4  & 71.2  & 80.5  & 68.6  & 92.8  & 97.6  & 55.7  & 85.0  & 91.0  \\
          & SAF   & 51.0  & 79.3  & 88.0  & 38.3  & 66.5  & 76.2  & 67.3  & 92.5  & 96.6  & 53.4  & 84.5  & 92.4  \\
    20\%  & SGR*  & 62.8  & 86.2  & 92.2  & 44.4  & 72.3  & 80.4  & 67.8  & 91.7  & 96.2  & 52.9  & 83.5  & 90.1  \\
          & SGR-C & 72.8  & 90.8  & 95.4  & 56.4  & 82.1  & 88.6  & 75.4  & 95.2  & 97.9  & 60.1  & 88.5  & 94.8  \\
          & NCR   & 75.0  & 93.9  & 97.5  & 58.3  & 83.0  & 89.0  & 77.7  & 95.5  & 98.2  & 62.5  & 89.3  & 95.3  \\
          & MSCN  & \textbf{\ 77.4 } & \textbf{\ 94.9 } & \textbf{\ 97.6 } & \textbf{\ 59.6 } & \textbf{\ 83.2 } & \textbf{\ 89.2 } & \textbf{\ 78.1 } & \textbf{\ 97.2 } & \textbf{\ 98.8 } & \textbf{\ 64.3 } & \textbf{\ 90.4 } & \textbf{\ 95.8 } \\
    \midrule
          & SCAN  & 27.7  & 57.6  & 68.8  & 16.2  & 39.3  & 49.8  & 40.8  & 73.5  & 84.9  & 5.4   & 15.1  & 21.0  \\
          & VSRN  & 14.3  & 37.6  & 50.0  & 12.1  & 30.0  & 39.4  & 23.5  & 54.7  & 69.3  & 16.0  & 47.8  & 65.9  \\
          & IMRAM & 9.1   & 26.6  & 38.2  & 2.7   & 8.4   & 12.7  & 21.3  & 60.2  & 75.9  & 22.3  & 52.8  & 64.3  \\
          & SAF   & 30.3  & 63.6  & 75.4  & 27.9  & 53.7  & 65.1  & 30.4  & 67.8  & 82.3  & 33.5  & 69.0  & 82.8  \\
    50\%  & SGR*  & 36.9  & 68.1  & 80.2  & 29.3  & 56.2  & 67.0  & 60.6  & 87.4  & 93.6  & 46.0  & 74.2  & 79.0  \\
          & SGR-C & 69.8  & 90.3  & 94.8  & 50.1  & 77.5  & 85.2  & 71.7  & 94.1  & 97.7  & 57.0  & 86.6  & 93.7  \\
          & NCR   & 72.9  & 93.0  & \textbf{\ 96.3 } & 54.3  & 79.8  & 86.5  & 74.6  & 94.6  & 97.8  & 59.1  & 87.8  & 94.5  \\
          & MSCN  & \textbf{\ 74.4 } & \textbf{\ 93.2 } & 96.0  & \textbf{\ 55.3 } & \textbf{\ 80.4 } & \textbf{\ 86.8 } & \textbf{\ 77.5 } & \textbf{\ 96.2 } & \textbf{\ 98.7 } & \textbf{\ 60.7 } & \textbf{\ 89.1 } & \textbf{\ 94.9 } \\
    \midrule
          & SCAN  & 5.6   & 19.3  & 27.4  & 2.2   & 8.0   & 12.8  & 18.1  & 43.1  & 57.4  & 0.3   & 1.3   & 2.3  \\
          & VSRN  & 0.8   & 2.5   & 4.1   & 0.5   & 1.5   & 2.7   & 5.1     & 15.7     & 24.6 & 2.5  & 8.8     & 13.3 \\
          & IMRAM & 1.3   & 3.1   & 3.9   & 0.3   & 1.2   & 2.8   & 7.1     & 20.0     & 33.4 & 5.3     & 15.2     & 22.0 \\
          & SAF   & 0.5   & 2.2   & 3.0   & 0.2   & 0.8   & 1.7   & 0.1   & 1.7   & 4.0   & 0.6   & 1.9   & 3.0  \\
    70\%  & SGR*  & 17.9  & 42.1  & 51.9  & 14.6  & 31.0  & 40.8  & 35.7  & 71.2  & 85.4  & 31.6  & 65.8  & 79.0  \\
          & SGR-C & 65.0  & 89.3  & \textbf{\ 94.7 } & 48.1  & \textbf{\ 74.5 } & \textbf{\ 81.1 } & 69.8  & 93.6  &  97.5  & 56.5  & 86.0  &  93.4  \\
          & NCR   & 16.1  & 38.5  & 52.8  & 11.0  & 29.5  & 41.4  & 35.4  & 69.5  & 83.4  & 31.5  & 66.4  & 81.1  \\
          & MSCN  & \textbf{\ 69.0 } & \textbf{\ 89.3 } & 93.0  & \textbf{\ 49.2 } & 73.1  & 79.0  & \textbf{\ 74.4 } & \textbf{\ 94.9 } & \textbf{97.7}  & \textbf{\ 58.8 } & \textbf{\ 87.2 } & \textbf{93.7}  \\
    \bottomrule
    \end{tabular}%
    }
  \caption{Image-text retrieval performance under synthetic noise rates of 20\%, 50\% and 70\% on Flickr30K and MS-COCO 1K, and the best results are highlighted in \textbf{bold}.}
  \label{tab:table2}%
\end{table*}%

\subsection{Comparison with the State-of-the-Art}
In this section, we conduct comparison experiments with 7 state-of-the-art methods that include SCAN \cite{lee2018stacked}, VSRN \cite{li2019visual}, IMRAM \cite{chen2020imram}, SGR, SAF, SGRAF \cite{diao2021similarity}, and NCR \cite{huang2021learning}. As Flick30K and MS-COCO are well annotated datasets, we generate synthetic noisy correspondence by randomly shuffling the training images and captions for a specific noise ratio, and we conduct experiments with three different level of noise ratios, \ie, 20\%, 50\%, and 70\%. For all methods, we choose the best checkpoint on the validation set and report its performance on the test set. Following NCR, we also report two strong baselines based on SGR, \ie, SGR-C and SGR*. In short, SGR-C uses only clean data for training, and SGR* employs a pre-training process while training without hard negatives to improve robustness. To evaluate our method on real-world noisy data, we conduct experiments on the CC152K, which is harvested from the Internet and contains about 3\% $ \sim $  20 \% noisy correspondence \cite{sharma2018conceptual}. 

\paragraph{Experiments on CC152K.} \cref{tab:table1} shows the results on CC152K with real noisy correspondence. On CC152K, our MSCN achieves the state-of-the-art performance in terms of all metrics. Specifically, the sum score of MSCN is 4.9 \% and 6.2 \% higher than the best baseline in text and image retrieval, respectively. In addition, we also present a more strict comparison (denoted by MSCN*) that trained our method with only 1,000 meta-data (approximately 0.67\% of all data). We can see that our MSCN* also achieves a competitive performance. Specifically, all metrics except R@1 of image retrieval are superior to the baselines.

\paragraph{Experiments on Flickr30K and MS-COCO.} \cref{tab:table2} shows the results on Flickr30K and MS-COCO under a range of synthetic noise rates. Following \cite{huang2021learning}, we report the results by averaging over 5 folds of 1K test images. From \cref{tab:table2}, we can draw the observation that noisy correspondence remarkably affect the performance of cross-modal retrieval methods. The retrieval accuracy will decrease fast with the noise rate increasing. Our MSCN is superior to the baselines in almost all metrics under different noise rates. In the low noise cases, \ie, 20\% and 50\%, our MSCN improves R@1 (sum of image and text) by 3.7\%, 2.2\%, 2.5\%, and 4.5\% in the four valuations compared with NCR. In the high noise case, \ie, 70\%, all baselines can easily overfit to noisy correspondence and results in poor performance. Our MSCN improves R@1 by 52.9\%, 38.2\%, 39.4\%, and 27.3\% in these four valuations compared with NCR. Even compared with the SGR-C trained only on clean data, our MSCN also achieves competitive performance. 

\begin{figure*}[h]
  \setlength {\belowcaptionskip} {-0.3cm}
  \centering
  \includegraphics[width=1\textwidth]{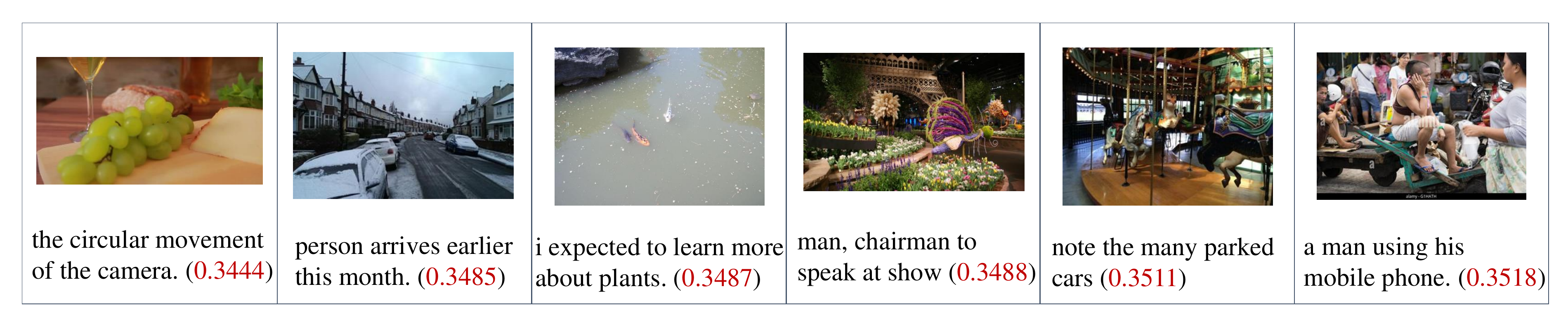}
  \caption{Real-world noisy examples detected by our MSCN.} 
  \label{fig:noisy samples}
\end{figure*}

\subsection{Ablation Study}

To evaluate the performance of the proposed components (\ie, self-adaptive margin and data purification), we conduct the ablation study on the Flickr30K with 20\% noisy correspondence. Note that for MSCN without $\mathcal{D}_{train}^{'}$, we only use a single model to perform the experiment. As shown in \cref{tab:table3}, we can see that the performance without $\hat{\gamma}$ or $\mathcal{D}_{train}^{\prime}$ are worse than the complete MSCN, which indicates that all components are important to achieve advantageous results. We explain more on the data selection strategy in the \emph{supplementary material}.

\begin{table}[htbp]
\setlength {\belowcaptionskip} {-0.2cm}
  \centering
  \resizebox{1\columnwidth}{!}{
     \renewcommand\arraystretch{.9}
    \small
    \begin{tabular}{ccc|ccc}
    \toprule
          & Method &       & \multicolumn{3}{c}{Image to Text} \\
    \midrule
    MSCN  &  w/o \small{$\hat{\gamma}$} & w/o \small{$\mathcal{D}_{train}^{\prime}$} & R@1   & R@5   & R@10 \\
    \midrule
    $\checkmark$ &       &       & \textbf{\ 77.4 } & \textbf{\ 94.9 } & \textbf{\ 97.6 } \\
    $\checkmark$ & $\checkmark$ &       & 75.3  & 94.5  & 97.2 \\
    $\checkmark$ &       & $\checkmark$ & 75.8  & 93.8  & 96.2 \\
    $\checkmark$ & $\checkmark$ & $\checkmark$ & 74.1  & 91.5  & 94.7 \\
    \midrule
          &       &       & \multicolumn{3}{c}{Text to Image} \\
    $\checkmark$ &       &       & \textbf{\ 59.6 } & \textbf{\ 83.2 } & \textbf{\ 89.2 } \\
    $\checkmark$ & $\checkmark$ &       & 58.3  & 83.1  & 88.9 \\
    $\checkmark$ &       & $\checkmark$ & 55.8  & 74.5  & 77.4 \\
    $\checkmark$ & $\checkmark$ & $\checkmark$ & 53.4  & 71.2  & 72.2 \\
    \bottomrule
    \end{tabular}%
    }
  \caption{Ablation studies on Flickr30K with 20\% noise rate.}
  \label{tab:table3}%
\end{table}%

\subsection{Progressive Comparison}

\begin{figure}[h]
\setlength {\belowcaptionskip} {-0.2cm}
\centering  
\includegraphics[width=0.75\columnwidth,trim=40 5 70 30,clip]{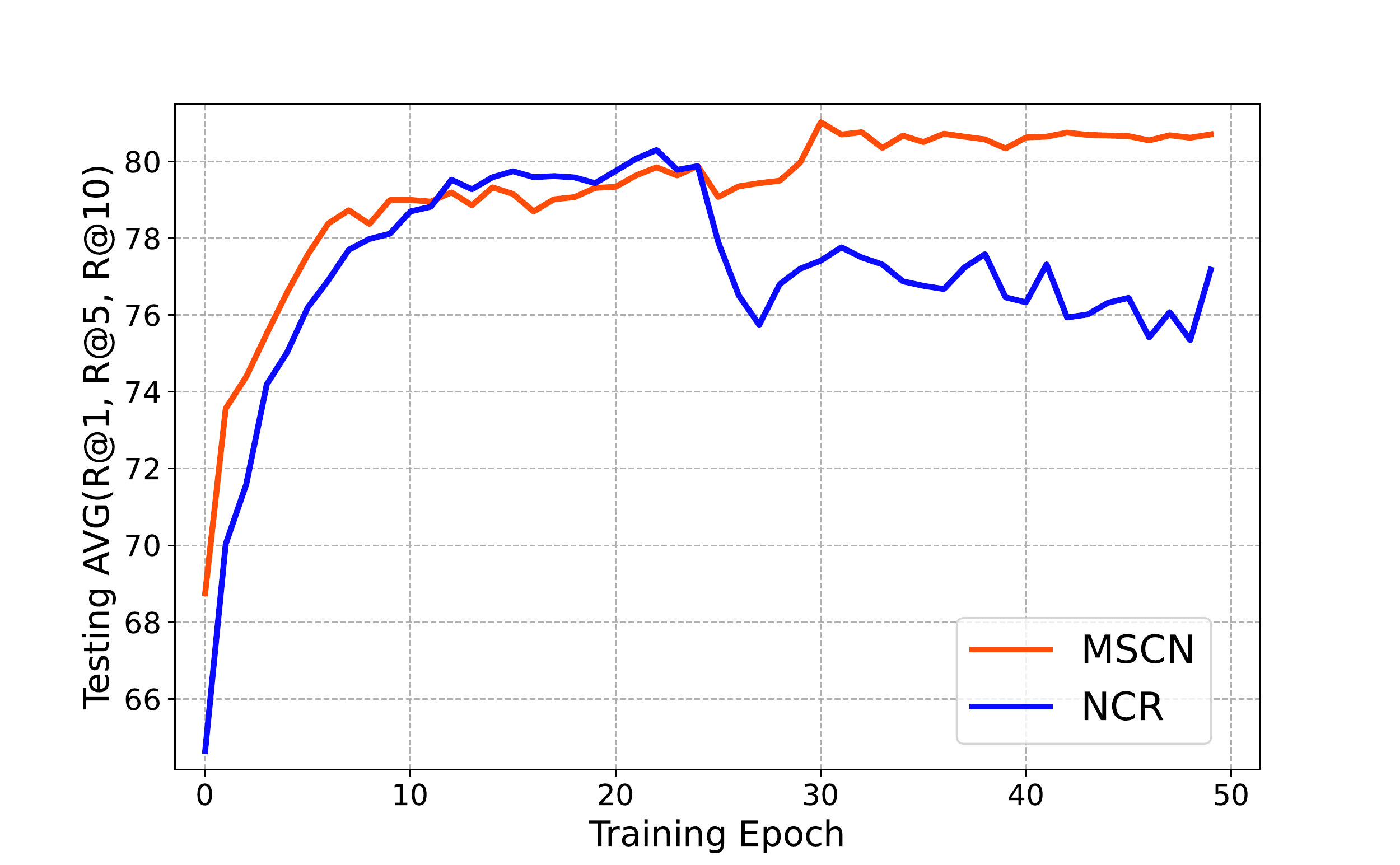}
\caption{Progressive performance comparison of MSCN and NCR on testing set as training proceeds.}
\label{acc}
\end{figure}

\cref{acc} plots the average of recalls on testing set of MSCN and NCR as training proceeds. We show a representative result using Flickr30K with 50\% noisy correspondence. From the result, we can see that our MSCN achieves a more stable performance in overall training process. Although NCR does not need an extra meta-set to guide the training, it relies on a clean validation set to choose the best model parameters. The stableness also indicates that our method has better alleviated the interference of noisy correspondence.

\subsection{Detected Noisy Samples}
\cref{fig:noisy samples} shows some real-world noisy image-text pairs detected (with low similarity score) by MSCN in Conceptual Captions. We also present the similarity score predicted by our MSCN. For most noisy pairs, the image and text contains completely irrelevant semantic information. Moreover, our MSCN can even find the subtle mistake in the noisy image-text pair. In the last pair, the picture shows a man covering his ears but wrongly caption as using a mobile phone. Although this picture is misleading, our method successfully detects the subtle difference.

\subsection{Visualization on Similarity Score}
\begin{figure}[h]
\setlength {\belowcaptionskip} {-0.3cm}
\centering
\subfloat[]{\includegraphics[width=0.5\columnwidth]{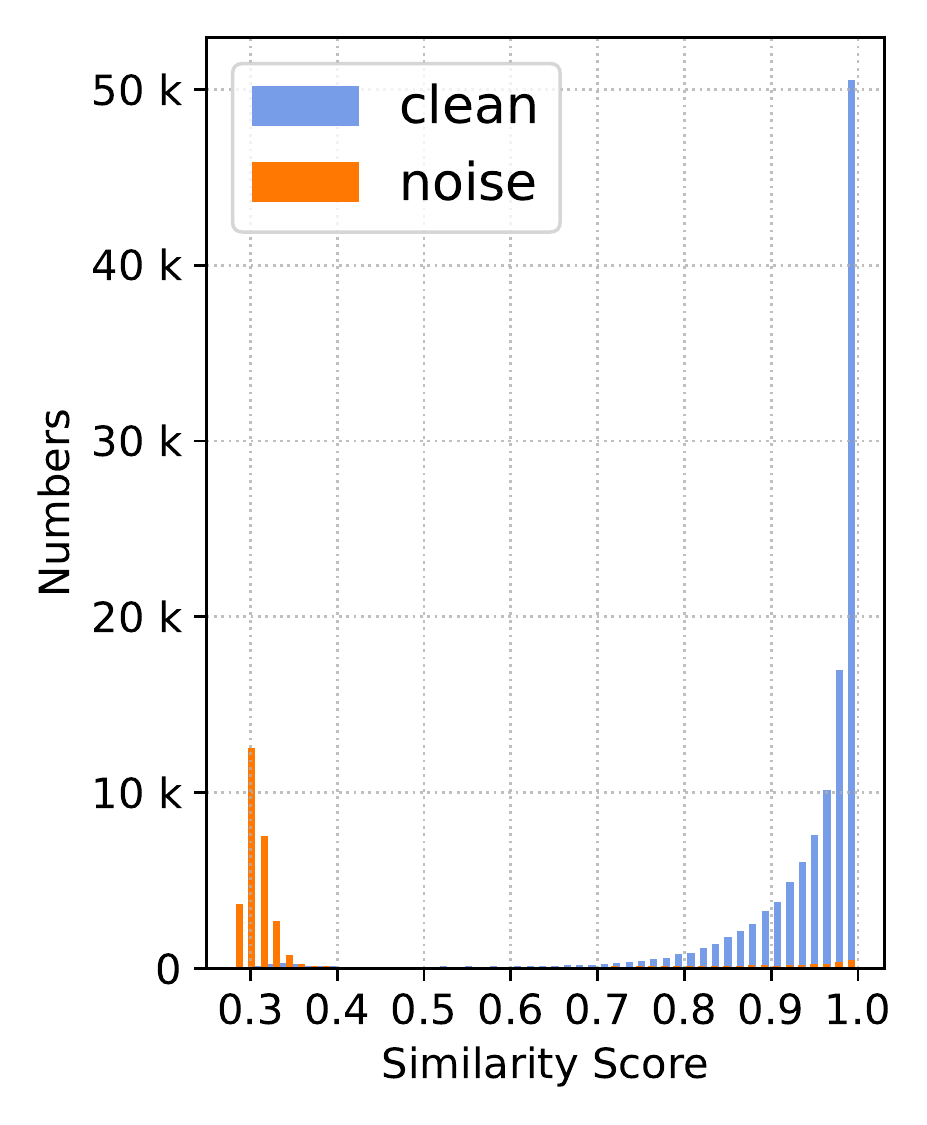}%
\label{0.2_visual}}
\subfloat[]{\includegraphics[width=0.5\columnwidth]{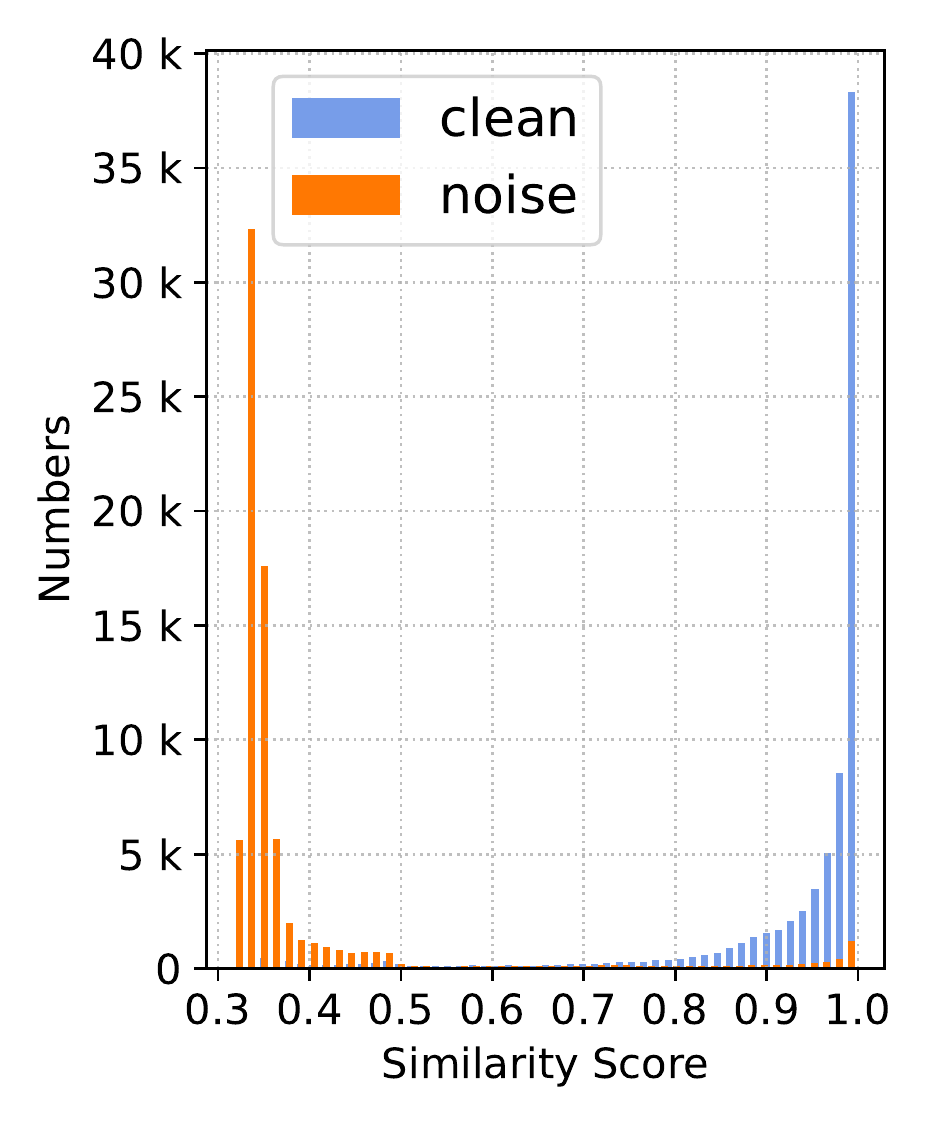}%
\label{0.5_visual}}
\caption{Per-sample similarity distribution on training data of Flickr30K. (a) 20\% noisy correspondence. (b) 50\% noisy correspondence.}
\label{visual}
\end{figure}

\cref{visual} plots the similarity score distribution predicted by MSCN for clean and noisy training samples. It can be seen that almost all large similarity scores belong to clean pairs, and the noisy pairs’ scores are smaller than clean samples, which implies that the trained MSCN can successfully provide corrected similarity scores.

\section{Conclusion}
In this paper, we explore a meta-learning method to address the problem of learning with noisy correspondence. Specifically, we propose a meta correction network (MSCN) to provide reliable similarity scores. Our MSCN is trained by a novel meta-process that views both positive and negative data as meta-knowledge to encourage MSCN to learn discrimination from them. To further mitigate noise interference, we design a data purification strategy that uses meta-data as prior knowledge to purify the noisy data efficiently. We conduct comprehensive experiments on three widely used datasets. The results validate the effectiveness of our method in both synthetic and real noises.
\vspace{-0.0cm}

\section*{Acknowledgements}
This work is supported by the National Key Research and Development Program of China (2020AAA0108800), National Nature Science Foundation of China (62192781, 62272374, 62202367, 62250009, 62137002, 61937001), Innovative Research Group of the National Natural Science Foundation of China (61721002), Innovation Research Team of Ministry of Education (IRT\_17R86), Project of China Knowledge Center for Engineering Science and Technology, and Project of Chinese academy of engineering ``The Online and Offline Mixed Educational Service System for `The Belt and Road' Training in MOOC China''. We would like to express our gratitude for the support of K. C. Wong Education Foundation.

{\small
\bibliographystyle{ieee_fullname}
\bibliography{egbib}
}

\end{document}